\documentclass{article}

\usepackage[english]{babel}

\usepackage[letterpaper,top=2cm,bottom=2cm,left=3cm,right=3cm,marginparwidth=1.75cm]{geometry}

\usepackage{amsmath}
\usepackage{graphicx}
\usepackage{booktabs}
\usepackage{array}
\usepackage{placeins} % if you want to use \FloatBarrier
\usepackage[colorlinks=true, allcolors=blue]{hyperref}
\usepackage[round]{natbib}
\usepackage{chngcntr}
\usepackage{appendix}
\usepackage{longtable}

\title{\textbf{BioMedLM: A 2.7B Parameter Language Model Trained On Biomedical Text}}
\author{Elliot Bolton¹\textsuperscript{\dag}, Abhinav Venigalla², Michihiro Yasunaga¹, David Hall¹, Betty Xiong¹, \and Tony Lee¹, Roxana Daneshjou¹, Jonathan Frankle², \and Percy Liang¹, Michael Carbin², and Christopher D. Manning¹\textsuperscript{\dag}  \\[1ex] ¹Stanford University\hspace*{2em} ²DataBricks \\[1ex]
\textsuperscript{\dag}Correspondence: elliotbolton@stanford.edu, manning@stanford.edu}
\date{March 2024}

\begin{document}
\maketitle

\begin{abstract}
\noindent Models such as GPT-4 and Med-PaLM 2 have demonstrated impressive performance on a wide variety of biomedical NLP tasks. 
However, these models have hundreds of billions of parameters, are computationally expensive to run, require users to send their input data over the internet, and are trained on unknown data sources. Can smaller, more targeted models compete? To address this question, we build and release BioMedLM, a 2.7 billion parameter GPT-style autoregressive model trained exclusively on PubMed abstracts and full articles. When fine-tuned, BioMedLM can produce strong multiple-choice biomedical question-answering results competitive with much larger models, such as achieving a score of 57.3\% on MedMCQA (dev) and 69.0\% on the MMLU Medical Genetics exam. BioMedLM can also be fine-tuned to produce useful answers to patient questions on medical topics. This demonstrates that smaller models can potentially serve as transparent, privacy-preserving, economical and environmentally friendly foundations for particular NLP applications, such as in biomedicine. The model is available on the Hugging Face Hub: \href{https://huggingface.co/stanford-crfm/BioMedLM}{https://huggingface.co/stanford-crfm/BioMedLM}.
\end{abstract}

\section{Introduction}

Large language models such as OpenAI's GPT-4 have become the dominant technology in modern natural language processing \citep{gpt4, zhao2023survey}. Trained on large corpora to predict the next token and refined with human feedback \citep{brown2020language, ouyang2022training, ziegler2020finetuning}, these models develop impressive capabilities in areas such as summarization and question-answering \citep{zhang2023benchmarking, goyal2023news, karpukhin2020dense}. While the focus has been on these models’ performance when responding to general English prompts, it is clear there is potential for specialist models to impact biomedical research and healthcare \citep{arora2023, shah2023, thirunavukaras2023}. Such applications include information retrieval and summarization from the ever-expanding biomedical literature \citep{wang2021, yang2020}, clinical information such as physician notes in electronic health records, and radiology reports \citep{murray2021, feblowitz2011, zhang2018learning}. Improving domain-specific language models will help accelerate biomedical discovery, drive down healthcare costs, and improve patient care.

% The world’s ever expanding biomedical knowledge is stored in millions of pages that no single human could hope to read in a lifetime. Doctors and hospital staff spend significant time engaged in basic information retrieval and presentation tasks such as querying clinical records for the results of a patient’s most recent lipid panel or summarizing a radiology report for non-technical readers. 

Large, general models like GPT-4 and Med-PaLM 2 have set new standards for performance on question-answering and information extraction \citep{usmle, medpalm, medpalm2}, but there are several drawbacks to these models.

They are costly to train and utilize. Compute for training and inference of large language models have increased 10- to 100-fold since 2015 \citep{sevilla2022compute}, translating to extremely high financial and environmental costs \citep{sharir2020cost, patterson2021carbon}. Organizations must pay expensive API fees which may be beyond their budget as well.

Due to their immense size and corporate secrecy, theses models must be run on giant compute clusters and accessed remotely over the internet. This eliminates the possibility of on-device inference and requires the transmission of sensitive data. The need to access these models via corporate APIs \citep{cobbe2023}, raises issues around third-party access of personal identifiable information (PII) \citep{lukas2023analyzing}. This conflicts with the medical need for patient data privacy, outlined by the Health Insurance Portability and Accountability Act (HIPAA) \citep{marks2023, mesko2023}.

The closed nature of these models create many complications as well. Any organization relying on these models is ultimately relying on a major tech company. If the tech company ceases to serve the model for any reason, an organization could immediately find any service it built on top of the model deactivated. 

One of the most prominent drawbacks of this closed nature is that the training data used for corporate models is a closely guarded secret. From a business perspective, this creates a risk to an organization using these models. For example, if an AI model provider has trained their model on data that violates copyright, the model may unpredictably become unavailable. From a practitioner perspective, the lack of transparency around model training data adds uncertainty to the quality and reliability of model responses. And from a research perspective, not knowing the constitution of model training data inhibits the ability to study the relationship between training data and downstream task performance.

Another consequence of the closed nature of these models is the inability to further fine-tune them for specific tasks. This is especially meaningful in the biomedical context, where models trained on general English data can benefit from further training on specialized tasks \citep{medpalm, medpalm2}.

In summary, these expenses and challenges can make language model technology inaccessible or nonviable to biomedical and/or healthcare organizations with limited resources and strict privacy requirements.

A promising direction is to build a smaller model and focus training on biomedical text, to achieve good domain-specific performance, and provide hope for a smaller solution. However, models like PubMedBERT \citep{pubmedbert}, SciBERT \citep{beltagy2019scibert} and BioBERT \citep{lee2019biobert} are much smaller by current standards. This presents an opportunity for a small-medium domain-specific model with strong performance. To further investigate the value of domain specialization, and the potential of smaller models, we introduce BioMedLM, a 2.7 billion parameter biomedical language model trained on PubMed abstracts and full articles, and made publicly available in December 2022. It is a GPT-2 style \citep{multitask} autoregressive model \citep{attention} with biomedical domain-specific tokenization, and trained exclusively on PubMed abstracts and full
articles \citep{gao2020pile}.

BioMedLM can achieve strong results on multiple-choice biomedical question answering, which are competitive with substantially larger models. This includes scores of 57.4\% on MedMCQA (dev) \citep{pal2022medmcqa} and 70.0\% on MMLU Medical Genetics \citep{hendrycks2021measuring}. Against a general-English model baseline, GPT-Neo 2.7B \citep{gao2020pile} (architecturally similar model to BioMedLM with a nearly identical parameter count), BioMedLM demonstrates an improvement in accuracy across three tasks: BioASQ \citep{bioasq}, PubMedQA \citep{pubmedqa} and MedQA \citep{medqa}. We also demonstrate question-answering requiring text generation capabilities; BioMedLM can produce useful multi-sentence answers to questions on medical topics from the HealthSearchQA question dataset \citep{medpalm}, such as ``What are the best ways to treat plantar fasciitis?".

BioMedLM is designed to address the drawbacks mentioned above. Its small size allow for it to be comfortably fine-tuned on a single A100 GPU and for inference to be run on a laptop. Organizations utilizing this model can serve it internally and never send private data out of their internal networks. Its training data is fully documented, allowing practitioners and researchers insight into model performance. And the model's open nature allows anyone to download it and fine-tune it as they see fit. 

As competition intensifies in AI, corporations are increasingly keeping the details of their models' architectures and training secret. Thus, it is vital for organizations to release models to the open source community so researchers can learn about the capabilities of language models and the trade-offs of different architecture and training design choices \citep{openfms}. We hope BioMedLM can spread knowledge about how to build LLMs and contribute to the body of knowledge around training domain-specific models and applying language models to biomedical NLP tasks. By releasing BioMedLM, we provide a performant smaller model which can be of use for biomedical tasks of language analysis and production.

\section{Related Work}

There are two broad types of models that have been deployed for biomedical question-answering NLP tasks. Large, general English models that have been adapted for the biomedical domain, like Med-PaLM 2 \citep{medpalm2}, are the state of the art on multiple-choice question-answering and multi-sentence answering. Small, domain-targeted models like DRAGON \citep{dragon} have shown impressive performance on multiple choice and text classification tasks for their size.

\paragraph{GPT-Neo 2.7B.}

GPT-Neo 2.7B is a GPT-style model with 2.7B parameters \citep{gpt-neo} trained on the Pile, a diverse English corpus 825 GB in size \citep{gao2020pile}. The Pile contains multiple sub-corpora, including general content from Common Crawl and specialized content like PubMed, Github, and FreeLaw. This model serves as an important baseline for comparison with our domain-specific models; because it is the same size as our model and has a similar architecture, we can better understand the impact of pre-training corpus composition on downstream task performance.

\paragraph{PubMedBERT.}

PubMedBERT is a BERT model trained on PubMed abstracts and articles that uses a biomedical-oriented tokenizer \citep{pubmedbert}. This model showed substantial improvement on biomedical NLP tasks over similar models trained on biomedical NLP tasks such as NER, relation extraction, and question-answering, demonstrating the potential of exclusively training on PubMed.

\paragraph{BioLinkBERT and DRAGON.}

BioLinkBERT and DRAGON are BERT-style models trained on PubMed \citep{linkbert, dragon}. They utilize augmented architectures and richer data sources to produce highly impressive performance on BLURB and biomedical question-answering \citep{pubmedbert}. BioLinkBERT exploits the link structure present in PubMed, while DRAGON builds on BioLinkBERT to leverage information from common biomedical knowledge graphs. With their high quality, domain-specific data, both models are able to compete with substantially larger models, despite only having 350-million parameters.

\paragraph{Galactica.}

Galactica is a 120 billion parameter language model  trained on scientific text that utilizes a tokenizer specialized for scientific text \citep{galactica}. It achieved impressive zero-shot performance on scientific tasks, including biomedical QA tasks, e.g., PubMedQA \citep{pubmedqa}, further demonstrating the value of training on specialized text and scale.

\paragraph{BioGPT.}

BioGPT is a GPT-style model trained on PubMed concurrent to this work \citep{biogpt}. This work shows the capabilities of a biomedical GPT-style model, including state-of-the-art results on PubMedQA when using artificial and unlabeled data and strong results on a variety of biomedical NLP tasks. While the BioGPT paper focuses more on relation extraction and summarization, we emphasize question-answering tasks.

\paragraph{Flan-PaLM and Med-PaLM.}

Flan-PaLM and Med-PaLM are fine-tuned versions of the 500 billion parameter PaLM language model \citep{palm, medpalm, medpalm2}. 
% which have shown state of the art results on standard medical question-answering tasks and impressive long-form question-answering capabilities. 
Flan-PaLM set the standard for MedQA and passed the USMLE without being specifically trained to perform on biomedical tasks, showing the capabilities of large-scale models trained on general language.

\paragraph{ChatGPT and GPT-4.}

\citet{usmle} evaluates ChatGPT (GPT-3.5) on a collection of publicly available USMLE questions and finds that ChatGPT can generally achieve passing scores without additional fine-tuning on specialized data. \citet{nori2023capabilities} assesses the capabilities of GPT-4 on USMLE and MultiMedQA benchmarks, and demonstrates that it exceeds performance of the previous GPT-3.5 and Med-PaLM. \citet{nori2023generalist} presents Medprompt, based on a composition of several prompting strategies, which elevates GPT-4 to better performance than Med-PaLM 2. While it is believed that scale can lead to significant progress in biomedical NLP, it is also difficult to assess train-test leakage in closed models.

\paragraph {Fully Open Language Models And Datasets.}

A thriving open source community is promoting full openness in model and dataset construction (i.e., training data, training process, and architecture are publicly known). Some of the first prominent models in this class were released by EleutherAI, including GPT-J \citep{gpt-j} and GPT-NeoX-20B \citep{black2022gptneox20b} which were both trained on the Pile. Together's RedPajama \citep{together2023redpajama} project, which aims to produce high quality open-source English datasets, has released 1 trillion and 30 trillion token datasets that have been used to train numerous open source models, including the 3 and 7 billion parameter INCITE models\citep{together2023incite}. Similarly the Allen Institute for AI has released OLMo \citep{groeneveld2024olmo} trained on the open Dolma \citep{soldaini2024dolma} training set.

\section{Model Design And Training}

\subsection{Model Architecture}

BioMedLM is an autoregressive, decoder-only Transformer \citep{attention} with 2.7 billion parameters. It is similar in architecture to GPT-2 \citep{multitask} with the settings in Table~\ref{tab:architecture-settings}.

\begin{table}[ht]
    \centering
    \begin{tabular}{l r}
     \toprule
     Parameter & Setting \\
     \midrule
     Hidden Size & 2560 \\
     %\hline
     Heads & 20 \\
     %\hline
     Layers & 32 \\
     %\hline
     Vocab Size & 28896 \\
     %\hline
     Sequence Length & 1024 \\
     \bottomrule
    \end{tabular}
    \caption{Model Architecture Settings}
    \label{tab:architecture-settings}
\end{table}
\FloatBarrier

The model uses learned absolute positional embeddings for each position in the sequence \citep{attention}.

\subsection{BioMedLM Tokenizer}

It has been shown across several domains that domain-specific tokenizers help \citep{sachidananda2021efficient}, including the biomedical domain \citep{pubmedbert, linkbert, dragon}, science in general \citep{galactica} and law \citep{legal}. Utilizing tokenizers customized for their domains allows information for important terms to be stored in corresponding embeddings, rather than split across multiple sub-term tokens.

% downstream task performance can be enhanced if a domain-specific tokenizer is used in concert with a domain-specific language model \citep{legal, sachidananda2021efficient}. PubMedBERT, BioLinkBERT, DRAGON, and Galactica are all examples of models that utilize specialized tokenizers for their domains of interest \citep{pubmedbert, linkbert, dragon, galactica}. In other specialized domains with specialized vocabularies like law \citep{legal}, utilizing tokenizers customized for these domains allows information for important terms to be stored in corresponding embeddings, rather than split across multiple sub-term tokens. 

BioMedLM uses a custom Byte-Pair Encoding (BPE) tokenizer \citep{bpe} that was trained on PubMed abstracts. The tokenizer was trained with the Hugging Face Tokenizers library \citep{tokenizerlib} with the settings in Table~\ref{tab:tokenizer-settings}.

\begin{table}[ht]
    \centering
    \begin{tabular}{l r}
     \toprule
     Parameter & Setting \\
     \midrule
     Vocab Size & 28896 \\
     %\hline
     Min Frequency & 2 \\
     %\hline
     Add Prefix Space & False \\
     %\hline
     Trim Offsets & True \\
     \bottomrule
    \end{tabular}
    \caption{Tokenizer Training Settings}
    \label{tab:tokenizer-settings}
\end{table}
\FloatBarrier

Using this custom tokenizer results in many common biomedical terms being represented by a single token. For instance, Table~\ref{tab:biomedical-tokenization} shows a sampling of biomedical terms that are split by the traditional GPT-2 tokenizer, but are not split by our BioMedLM tokenizer.

\begin{table}[ht]
    \centering
    \begin{tabular}{l l}
    \toprule
     BioMedLM Tokenization & GPT2 Tokenization\\
     \midrule
     chromatography & chrom/atography \\
     cytotoxicity & cyt/ot/oxicity \\
     ECG & EC/G \\
     GATA & G/ATA \\
     Immunohistochemistry & Imm/un/oh/ist/ochemistry \\
     myocardium & my/ocard/ium \\
     nanoparticles & nanop/articles \\
     photosynthesis & photos/ynthesis \\
     probiotic & probiotic \\
     thrombin & th/rom/bin \\
     \bottomrule
    \end{tabular}
    \caption{Examples of BioMedLM and GPT2 Tokenizations of Biomedical Terms}
    \label{tab:biomedical-tokenization}
\end{table}
\FloatBarrier

Consider a term like “thrombin”. The standard Hugging Face GPT-2 tokenizer splits this token into “th”, “rom”, and “bin”. With the standard tokenizer, information about thrombin is spread across these three tokens and shared with completely unrelated words that have small sub-word overlap. Compounding the issue, these sub-word tokens do not  correspond to meaningful units the way tokens such as “bio” or “photo” might. The BioMedLM tokenizer preserves this word as a single token, without storing information in disparate tokens.

This improved treatment of domain-specific terminology translates to improved downstream task performance. We ran multiple experiments at the 125 million parameter scale to test out possible design choices. Experiments with different tokenizers but otherwise identical settings showed a meaningful improvement on the MedQA task over 5 random seeds (see Table~\ref{tab:tokenizer-impact}).

\begin{table}[ht]
    \centering
    \begin{tabular}{l r}
     \toprule
     Model & MedQA Test Accuracy \\
     \midrule
     BioMedLM-125m (GPT-2 tokenizer) & 33.05 \\
     BioMedLM-125m (BioMedLM tokenizer) & 34.98 \\
     \bottomrule
    \end{tabular}
    \caption{Tokenizer Impact on MedQA Performance}
    \label{tab:tokenizer-impact}
\end{table}
\FloatBarrier

\subsection{Pre-training}

BioMedLM was trained on the subparts of The Pile (as of November 2022) containing PubMed abstracts and full articles \citep{gao2020pile}. There were 34.6 billion tokens in the training corpus. The training run performed 8.67 passes through the data.

The model was trained on MosaicML Cloud, a platform designed for large workloads like training LLMs. The code used during pre-training utilized Flash Attention \citep{flashattention}, an algorithm for accelerating and reducing the memory requirements of calculating attention. This was a crucial implementation detail that allowed us to pre-train a model of BioMedLM’s scale on our available compute resources. We used Hugging Face’s GPT-2 model code \citep{multitask}, the Composer training library \citep{mosaicml2022composer} and PyTorch FSDP \citep{paszke2019pytorch}, employing multi-node training across 128 40GB Nvidia A100 GPUs. The total training run was completed in 6.25 days.

The model was trained with batch size 1024 and sequence length 1024 ($\sim$1 million tokens per batch) for 300 billion tokens using Decoupled AdamW \citep{decoupled}  to minimize the standard cross entropy loss of the subsequent token. The settings used for training are summarized below in Table~\ref{tab:pretraining-settings}.

\begin{table}[ht]
    \centering
    \begin{tabular}{l r}
     \toprule
     Parameter & Setting \\
     \midrule
     Tokens Per Batch & 1048576 \\
     Learning Rate & 1.6e-4 \\
     Scheduler & Cosine w/linear warmup (100 batches)\\
     Epsilon & 1e-8 \\
     Betas & [0.9, 0.95] \\
     Weight Decay & 1.6e-5 \\
     \bottomrule
    \end{tabular}
    \caption{Pre-training Settings}
    \label{tab:pretraining-settings}
\end{table}
\FloatBarrier

Initial trial runs at the 1.5 billion scale suffered from consistent training loss divergences. These runs used fp16/fp32 mixed precision (using NVIDIA's AMP library) similar to the Mistral project \citep{bolton2021mistral}, generally using fp16 precision but upcasting to fp32 when computing attention. The divergence issues were resolved when we switched to using bf16 \citep{google-bfloat16}. 

The final training process of the 2.7 billion parameter model did not suffer from any divergences. During the final run, parameters and optimizer states were stored in fp32, gradient communication was done in fp32, and training was done in bf16 (see Table~\ref{tab:precision-settings}).

\begin{table}[ht]
    \centering
    \begin{tabular}{l r l}
     \toprule
     Parameter & Library & Setting \\
     \midrule
     Compute Precision & Composer & bf16 \\
     Parameter Storage & FSDP & fp32 \\
     Optimizer Storage & FSDP & fp32 \\
     Gradient Communication & FSDP & fp32\\
     \bottomrule
    \end{tabular}
    \caption{Mixed Precision Settings}
    \label{tab:precision-settings}
\end{table}
\FloatBarrier

As we were preparing the training run, we were unsure of the benefits of training out to 300B tokens for language model perplexity and downstream task performance. While most models of this scale (e.g., GPT-Neo 2.7B \citep{gao2020pile}) are trained to 300-400B tokens, the datasets those models use are vastly larger than PubMed. For instance, The Pile is 8x the size of its PubMed subcorpora.

\begin{figure}[ht]
    \centering
    {{\includegraphics[width=7cm]{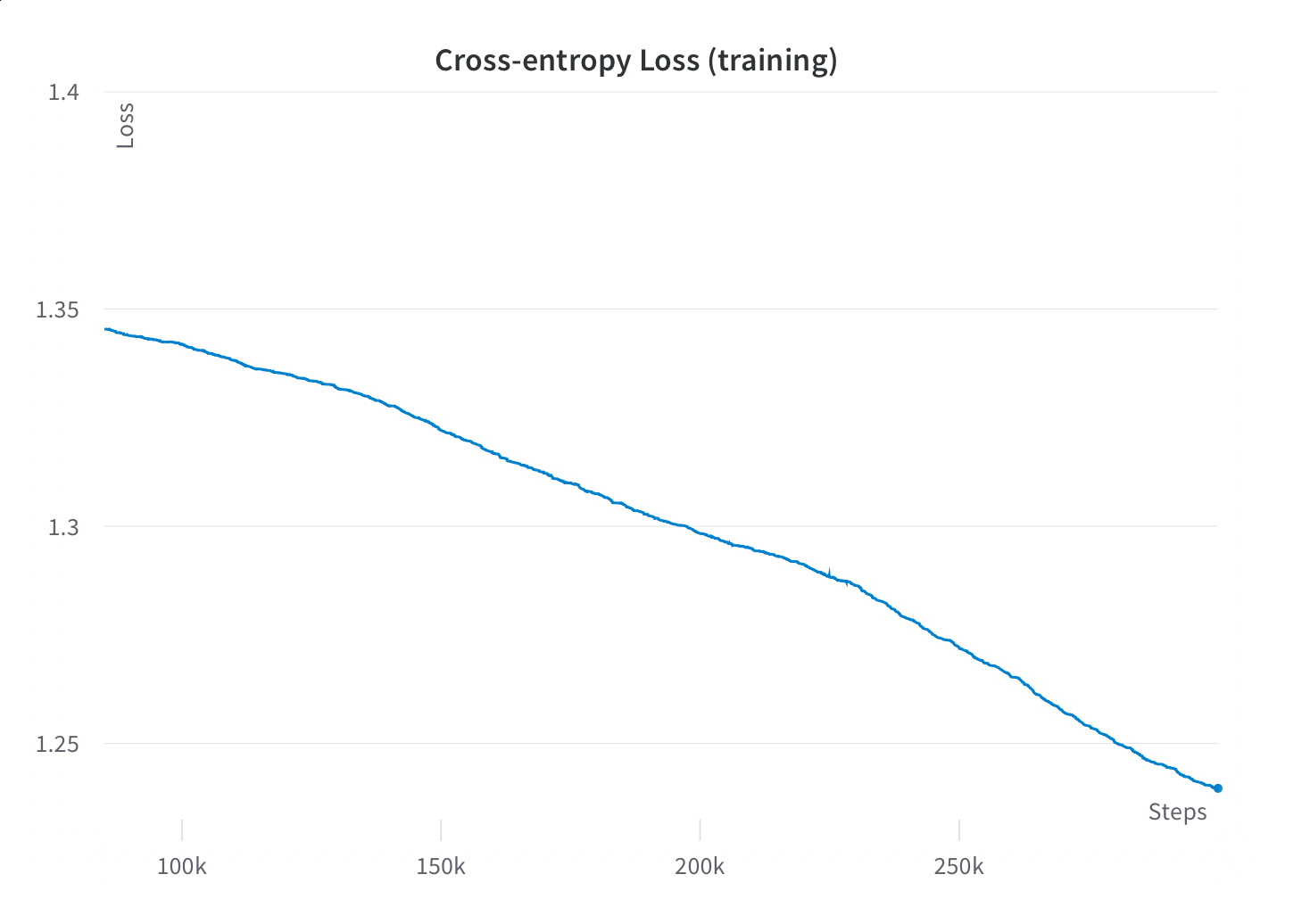} }}%
    \qquad
    {{\includegraphics[width=7cm]{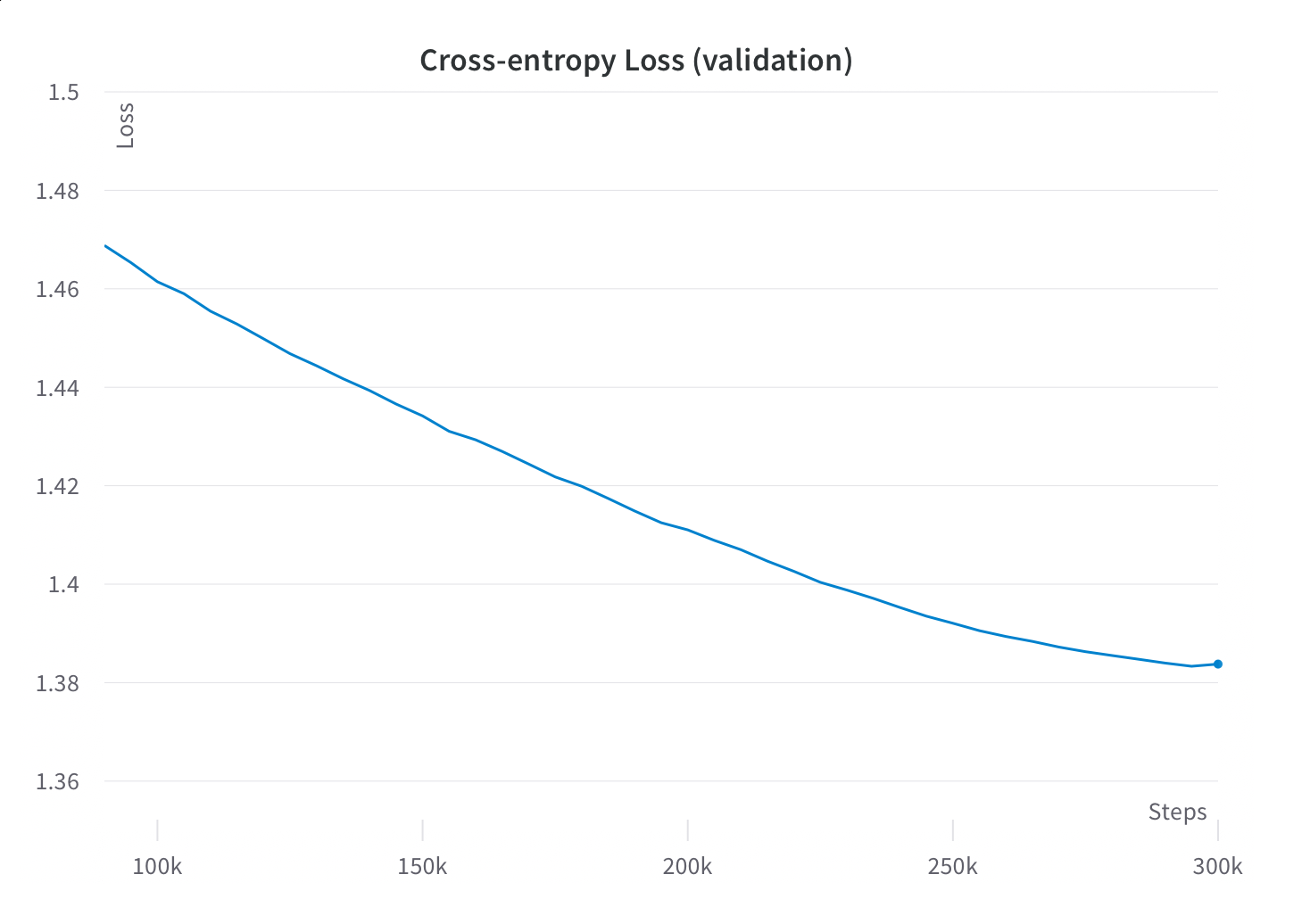} }}%
    \caption{Train and Validation Loss after 100k Batches}%
    \label{fig:train-val-loss}%
\end{figure}

Fortunately, we did continue to see steady perplexity improvements on the validation and training sets for the entirety of training (Figure~\ref{fig:train-val-loss}), and preliminary experiments showed improved downstream task performance as we trained out to the full 300B tokens. Consistent with concurrent work showing that long training is useful for small models \citep{devries2023chinchilla_analysis, touvron2023llama}, our takeaway is that it was indeed worth it to train for the full 300B tokens, despite having dramatically more passes through the data than comparable models \citep[e.g.,][]{bolton2021mistral}.

\subsection{Fine-tuning}

\paragraph{Fine-tuning for Question-Answering Tasks.}

We fine-tune the pre-trained BioMedLM for downstream question-answering tasks. Different tasks have different output formats. For MedMCQA and MedQA, the prompt contains a passage that ends with a question, from which the model is asked to pick the correct option. For MMLU, we fine-tuned a multiple-choice model on the MedMCQA and MedQA training data and evaluated it on the selection of MMLU exams related to biomedicine used in previous work \citep{medpalm}.

For generative models, it is common to supply the question text with answer options as the input context and record the generated answer from the model's response. The model's performance can be improved by including few-shot examples in the input context or directly fine-tuning the model to respond to multiple choice questions with the appropriate answer.

Our experiments demonstrated multiple accuracy points of improvement by using an architecture specialized for multiple-choice exams. For each question, the question context is concatenated with each answer option. Each question context and ending is run through the transformer, and the hidden state at the end of the full sequence with the proposed ending is run through a linear classifier which creates scores for each ending which are used to select the answer. This multiple-choice-specific architecture was used for our MedQA, MedMCQA, and MMLU results.

For PubMedQA and BioASQ, the context and question were concatenated, and the hidden state for a special token at the end of the concatenated sequence was fed to a linear classifier which produced scores for ``yes/no/maybe'' or ``yes/no''. Effectively the questions were treated as sequence classification examples for these two tasks. A small set of experiments directly training BioMedLM to generate ``yes'', ``no'', or ``maybe'' for PubMedQA yielded similar results, so it is not clear if the sequence classification architecture provided any performance boost.

Examples for each question type can be found in the supplementary material.

Standard Hugging Face code for these specializations can be found on the GitHub accompanying this paper: \url{https://github.com/stanford-crfm/BioMedLM}.

\paragraph{Optimal Format for PubMedQA and BioASQ.}

Since BioMedLM is a unidirectional model, it requires extra care during fine-tuning for the BioASQ and PubMedQA tasks. Unlike MedQA which simply contains a passage that ends with a question, BioASQ and PubMedQA come in a structured format that separates the question and context passage, leaving it to the user to arrange these components into prompts for the language model. In our initial experiments we placed the question at the beginning and followed it with the context. We were able to see significant performance gains on these tasks when we introduced special tokens and reorganized the components into the following format:

[Context Token]  “Text of context …”  [Question Token]  “Text of question …”  [Answer Token]

\paragraph{Fine-tuning for Long-Form Question-Answer Generation.}

To evaluate our model's generative capabilities, we fine-tuned it on a collection of medical knowledge question/answer pairs directed towards patients. First, we collected over 53,000 questions from publicly available sources on the web. A typical example would be a question like ``What are the best ways to treat plantar fasciitis?" and the corresponding answer a medical doctor would give.

\section{Biomedical Question-Answering Results}

With a trained model in hand, we began exploring its capabilities in the biomedical QA space. We focused on 5 standard biomedical NLP QA tasks: MedMCQA \citep{pal2022medmcqa}, MedQA \citep{medqa}, MMLU \citep{hendrycks2021measuring}, PubMedQA \citep{pubmedqa}, and BioASQ \citep{bioasq}, as well as summarization of consumer health questions \citep{medqasum}. Our model was able to produce results competitive with substantially larger models, and was often able to surpass the bi-directional models with augmented architectures trained on richer biomedical data. Examples of all these question-answering datasets are shown in the Appendix.

\subsection{MedMCQA}

This dataset contains 182822/4183/6150 train/dev/test questions drawn from AIIMS PG and NEET PG questions found on the web and in books \citep{pal2022medmcqa}. There is a wide variety of questions covering topics ranging from clinical questions to fundamental biochemistry, with each question having four multiple-choice options. Results are presented in  Table~\ref{tab:medmcqa}.

\begin{table}[ht]
    \centering
    \begin{tabular}{l r l r}
     \toprule
     & & & MedMCQA Test \\
     Model & Params & Method & Accuracy \\
     \midrule
     GPT-4 & -- & few-shot & 72.4 \\
     Flan-PaLM & 540B & few-shot & 57.6 \\
     BioMedLM & 2.7B & fine-tune & 57.3 \\
     Galactica & 120B & zero-shot & 52.9 \\
     GPT-3.5 & 175B & few-shot & 51.0 \\
     \bottomrule
    \end{tabular}
    \caption{MedMCQA Performance of Various Systems
    }
    \label{tab:medmcqa}
\end{table}
\FloatBarrier

% Despite being 200x smaller, BioMedLM is able to generally match the performance of Flan-PaLM on this task. The clear factor for this task is the substantially larger training data set size. This suggests that drastically smaller models can obtain competitive results with larger models when given sufficient training examples on a specific task such as medical question-answering.
% 
\subsection{MedQA}

This dataset contains 10178/1272/1273  train/dev/test questions drawn from USMLE questions found on the web \citep{usmle}. A standard question presents a medical scenario that a physician should be able to answer, and four options. Results are presented in  Table~\ref{tab:medqa}.

\begin{table}[ht]
    \centering
    \begin{tabular}{l r l l r}
     \toprule
      & & Model & & MedQA Test \\
     Model & Params & Openness & Method & Accuracy \\
     \midrule
     Med-PaLM 2 & -- & closed & few-shot & 85.4 \\
     GPT-4 & -- & closed & few-shot & 81.4 \\
     Flan-PaLM & 540B & closed & few-shot & 67.2 \\
     BioMedLM (MedMCQA data + classifier) & 2.7B & fully open & fine-tune & 54.7 \\
     GPT-3.5 & 175B & closed & few-shot & 53.6 \\
     BioMedLM (classifier) & 2.7B & fully open & fine-tune & 50.3 \\
     DRAGON & 360M & fully open & fine-tune & 47.5 \\
     BioLinkBERT & 340M & fully open & fine-tune & 45.1 \\
     Galactica & 120B & open weights & zero-shot & 44.4 \\
     GPT-Neo 2.7B & 2.7B & fully open & fine-tune & 37.7 \\
     \bottomrule
    \end{tabular}
    \caption{MedQA Performance of Various Systems}\label{tab:medqa}
\end{table}
\FloatBarrier

% At this point systems with hundreds of billions of parameters have become dominant on this task and are rivaling expert human performance. But BioMedLM can perform reasonably well for its size and even outperform the substantially larger Galactica.
% 
\subsection{MMLU}

The MMLU collection of questions covers a wide variety of subject areas, from biology and mathematics to history and philosophy \citep{hendrycks2021measuring}. The tests also range over different academic levels, from high school to professional. Several of the tests are relevant to the biomedical domain. Results are presented in  Table~\ref{tab:mmlu}.

\begin{table}[ht]
    \centering
    \begin{tabular}{l r l r r r r}
     \toprule
     Model & Params & Method & Clinical & Professional & College & Medical \\
      & & & Knowledge & Medicine & Biology & Genetics \\
     \midrule
     GPT-4 & -- & few-shot & 86.4 & 93.8 & 93.8 & 92.0 \\
     Flan-PaLM & 540B & few-shot & 80.4 & 83.8 & 88.9 & 75.0 \\
     GPT 3.5 & 175B & zero-shot & 69.8 & 70.2 & 72.2 & 70.0 \\
     Galactica & 120B & zero-shot & 59.2 & 59.6 & 68.8 & 70.0 \\
     BioMedLM & 2.7B & fine-tune & 59.6 & 63.1 & 60.7 & 69.0 \\
     \bottomrule
    \end{tabular}
    \caption{MMLU Performance of Various Systems
    }
    \label{tab:mmlu}
\end{table}
\FloatBarrier

% Despite only being 2.7 billion parameters in size, BioMedLM is able to come close to matching Flan-PaLM on the medical genetics topic, and can generally match Galactica on topics with medical emphasis. It clearly struggles on the College Biology topic, which is likely due to the domain drift between college level biology topics and formal PubMed articles.
% 
\subsection{PubMedQA}

This dataset contains 450/50/500 train/dev/test questions constructed from combining example PubMed article abstracts with questions derived by altering the paper title and answers related to the abstract \citep{pubmedqa}. Each question can be answered yes/no/maybe. Results are presented in  Table~\ref{tab:pubmedqa}.

\begin{table}[ht]
    \centering
    \begin{tabular}{l r l r}
     \toprule
     & & & PubMedQA \\
     Model & Params & Method &  Test Accuracy \\
     \midrule
     BioGPT (w/extra data) & 1.5B & fine-tune & 81.0 \\ 
     Flan-PaLM & 540B & few-shot & 79.0 \\
     Galactica & 120B & zero-shot & 77.6 \\
     GPT-4 & - & zero-shot & 75.20 \\
     BioMedLM & 2.7B & fine-tune & \textbf{74.4} \\
     DRAGON & 360M & fine-tune & 73.4 \\
     BioLinkBERT & 340M & fine-tune & 72.4 \\
     GPT-Neo 2.7B & 2.7B & fine-tune & 66.1 \\
     \bottomrule
    \end{tabular}
    \caption{PubMedQA Performance of Various Systems}\label{tab:pubmedqa}
\end{table}
\FloatBarrier

% Once again BioMedLM is able to get within striking distance of substantially larger models. Due to the small training set size of PubMedQA, our best system fine-tuned for this task is further away despite BioMedLM being trained on PubMed articles. Recently BioGPT has shown great results by being fine-tuned on the extra noisy label and unlabeled data provided with PubMedQA, and we feel a promising future direction would be exploring the 2-phase fine-tuning approach to further improve BioMedLM's performance. 
% 
\subsection{BioASQ}

This task is similar to PubMedQA, containing 670/75/140 train/dev/test questions constructed from combining example biomedical passages and questions relevant to the passage \citep{bioasq}. Each question can be answered yes/no. Results are presented in  Table~\ref{tab:bioasq}.

\begin{table}[ht]
    \centering
    \begin{tabular}{l r l r}
     \toprule
     & & & BioASQ \\
     Model & Params & Method &  Test Accuracy \\
     \midrule
     DRAGON & 360M & fine-tune & 96.4 \\ 
     BioMedLM & 2.7B & fine-tune & 95.7 \\
     BioLinkBERT & 340M & fine-tune & 94.9 \\
     Galactica & 120B & zero-shot & 94.3 \\
     GPT-Neo 2.7B & 2.7B & fine-tune & 67.1 \\
     \bottomrule
    \end{tabular}
    \caption{BioASQ Performance of Various Systems}\label{tab:bioasq}
\end{table}
\FloatBarrier

% BioMedLM is able to come close to the state of the art on this question-answering task as well and out perform BioLinkBERT and Galactica. We speculate it has worse relative performance on this task due to the longer lengths of BioASQ questions.
% 
\section{Free Response Question Answering}

While multiple-choice question-answering can demonstrate a model's medical knowledge, it is not straight forward to turn a multiple-choice answering system into a practical tool. Ideally, one should be able to direct medical questions to the model and get back multi-sentence generated answers, as if consulting an expert colleague. 

To test our model's ability to produce paragraph level responses, we fine-tuned it on health question answer pairs derived from publicly available data on the web. We constructed a training set based on FAQ pages and converting subsections of Wikipedia into a question answer format. Content from FAQ pages was directly added to the training set, with each question on the page being added as one training example. Wikipedia articles were used to generate more examples. For instance, the ``Signs and symptom'' section of the ``Plantar fasciitis'' Wikipedia page was converted to the question ``What are the signs and symptoms of plantar fasciitis?'' with the corresponding section text used as the answer. Thousands more question answer pairs were derived from Wikipedia in this manner.

It is important to note that BioMedLM needs to be fine-tuned on an instruction data set like this to enable it to answer questions like this. The base BioMedLM model was trained to replicate text in PubMed articles, so it is not well suited for question response.

In Table \ref{tab:consumerqa}, we demonstrate our system's performance on some example questions. We are planning a more formal evaluation of generation quality in future work.

\begin{table*}[p]
    \centering
    \small
    \begin{tabular}{p{3.75cm} p{11cm}}
     \toprule
     Question & BioMedLM Answer \\
     \midrule
     What does vitamin D do? & {Vitamin D plays an important role in maintaining your bones and supporting your immune system. It's also involved in many other body processes. 
     
     Vitamin D is a fat-soluble vitamin. Your body needs to get this vitamin from your diet and/or your sun exposure. You can get vitamin D from foods that contain vitamin D, such as fatty fish, eggs and fortified milk. You can also get vitamin D from sun exposure. 
     
    Your body needs to get enough vitamin D to function properly. Vitamin D deficiency can cause rickets in children and osteomalacia in adults. Vitamin D deficiency can also cause muscle weakness and an increased risk of falls and bone fractures.
    
    Vitamin D deficiency is a common condition. In the United States, about 1 in 10 people have vitamin D deficiency.}\\
    \midrule
    What are good treatments for arthritis? & {There are many treatments for arthritis. Your healthcare provider will recommend the treatment that's right for you based on your symptoms and the type of arthritis you have. Treatments for arthritis include:
    
    Medications: Your healthcare provider may prescribe medications to reduce pain and inflammation. Common medications include:
    
    Nonsteroidal anti-inflammatory drugs (NSAIDs).
    
    Corticosteroids.
    
    Pain relievers.
    
    Physical therapy: Physical therapy can help you manage arthritis symptoms. Your healthcare provider may recommend exercises to strengthen muscles and improve flexibility.
    
    Surgery: You may need surgery if your arthritis is severe or you don't respond to other treatments. Your healthcare provider may recommend joint replacement surgery or osteotomy.
    
    Complications of arthritis treatments
    You may experience side effects from arthritis treatments. Your healthcare provider will discuss the benefits and risks of your treatment options with you.} \\
    \midrule
    What are the common causes of lower back pain? & {The most common causes of lower back pain include:

    Degenerative disk disease: Damage to the disks in your lower back causes your disks to wear down. This wear and tear can lead to lower back pain.
    
    Spinal stenosis: A narrowing of your spinal canal can cause pressure on your nerves and lower back pain.
    
    Scoliosis: A curved spine can cause pain in your lower back.
    
    Herniated disk: A bulge in one of your disks causes pain.
    
    Spinal infection: Infections in your spine can cause pain and other symptoms.
    
    Pregnancy: Your body is more sensitive during pregnancy, which can cause back pain.
    
    Trauma: A fall, car accident or other injury can cause pain in your lower back.} \\
    \midrule
    What are the best ways to unclog your ears? & {There are several ways to unclog your ears. Your healthcare provider will recommend the method that's right for you.
    
    Ear drops: Your healthcare provider will apply ear drops to your ears. The drops may contain a steroid or antibiotic.
    
    Ear tubes: Your healthcare provider will surgically insert a small tube into your ear canal. The tube drains fluid and helps drain ear infections.
    
    Surgery: Your healthcare provider may perform ear surgery to remove excess ear tissue or open up your ear canal.} \\
    \midrule
    % What do night sweats indicate? & {Night sweats are a symptom of many different conditions. If you have night sweats, it's important to talk to a healthcare provider about what's causing them.} \\
    % \midrule
    % How serious is hepatitis A? & {Hepatitis A is usually mild and doesn't cause long-term health problems. Most people recover fully from the infection without any medical treatment.} \\
    % \midrule
    How can I stop diarrhea? & {The best way to stop diarrhea is to treat the cause. If you have an infection, your healthcare provider may prescribe antibiotics. If you have a food poisoning outbreak, you can go to the emergency room or call your healthcare provider to get help. If you have chronic diarrhea, you may need to change your diet. If you have an autoimmune disease, your healthcare provider may prescribe medication. If you take a medication that causes diarrhea, you may need to stop taking it.} \\
    % \midrule
    % What happens if listeria is left untreated? & {Listeria can cause serious illness and death if it's not treated. Up to 30\% of people with listeriosis die from the infection.} \\
    % \midrule
    % Does stress cause nosebleed? & {Yes. Stress, anxiety and panic can increase your heart rate and blood pressure, which can lead to nosebleeds.} \\
    \bottomrule
    \end{tabular}
    \caption{BioMedLM Example Responses To Consumer Health Questions}\label{tab:consumerqa}
\end{table*}

% The model's responses to medical questions are generally accurate. At times the model is clearly delivering vague answers and referring the patient to a medical professional. But several questions get very detailed answers.

% One issue that the model's generations suffer from is hallucination, especially around numerical values. Both numerical examples in the questions above appear to be wrong. A brief search seems to indicate that 20\% of patients die of listeriosis rather than the 30\% suggested by the model's answer. And it appears that 35\% of people in the US have Vitamin D deficiency rather than 10\%. A complete application to provide answers to patients would need additional safeguards to correct incorrect numerical values such as these examples. But BioMedLM does have the potential to be a component of such an application.
% 
\section{Analysis}

\subsection{BioMedLM Performance Across Different Biomedical Tasks}

At this point, systems with hundreds of billions of parameters have become dominant on some medical tasks (e.g. MedQA), and are rivaling expert human performance. However, in all question-answer tasks (MedMCQA, MedQA, MMLU, PubMedQA and BioASQ), BioMedLM is able to achieve comparable results to larger models. 

For MedMCQA, the substantially larger training data set size of 182822 allows BioMedLM, despite being 200x smaller, to match the performance of Flan-PaLM (57.38 vs.\ 57.6). For MedQA, BioMedLM (54.1) even outperforms the larger Galactica model (44.4) by a significant margin. For MMLU, BioMedLM is able to come close to matching Flan-PaLM on the Medical Genetics topic (70.0 and 75.0, respectively), and can generally match Galactica on topics with medical emphasis, i.e., Professional Medicine (61.76 and 59.6, respectively). It performs worst on the College Biology topic (57.64), which is likely due to the domain drift between college level biology topics and formal PubMed articles. 

For PubMedQA, due to the small training set size, our best system fine-tuned for this task is further away, despite BioMedLM (74.4) being trained on PubMed articles. Recently BioGPT (81.0) has shown great results by being fine-tuned on the extra noisy label and unlabeled data provided with PubMedQA, and we feel a promising future direction would be exploring their 2-phase fine-tuning approach to further improve BioMedLM's performance. In BioASQ, BioMedLM (95.7) outperforms BioLinkBERT (94.9) and Galactica (94.3).

\subsection{Question-Answering with Text Generation} 

The model's responses to medical questions are generally accurate. At times the model is clearly delivering vague answers and referring the patient to a medical professional. It is worth noting that several questions get very detailed answers.

One issue that persists is hallucination, especially around numerical values. Both numerical examples in Table \ref{tab:consumerqa} appear to be wrong. Basic fact-checking reveals that 35\% of people in the US have Vitamin D deficiency rather than 10\%. A complete application to provide answers to patients would need additional safeguards to correct incorrect numerical values such as these examples. However, as of now, BioMedLM does have the potential to be a component of such an application.

\subsection{Comparison with GPT-Neo 2.7B Baseline}

\begin{figure}[ht]
\centering
\includegraphics[width=0.7\textwidth]{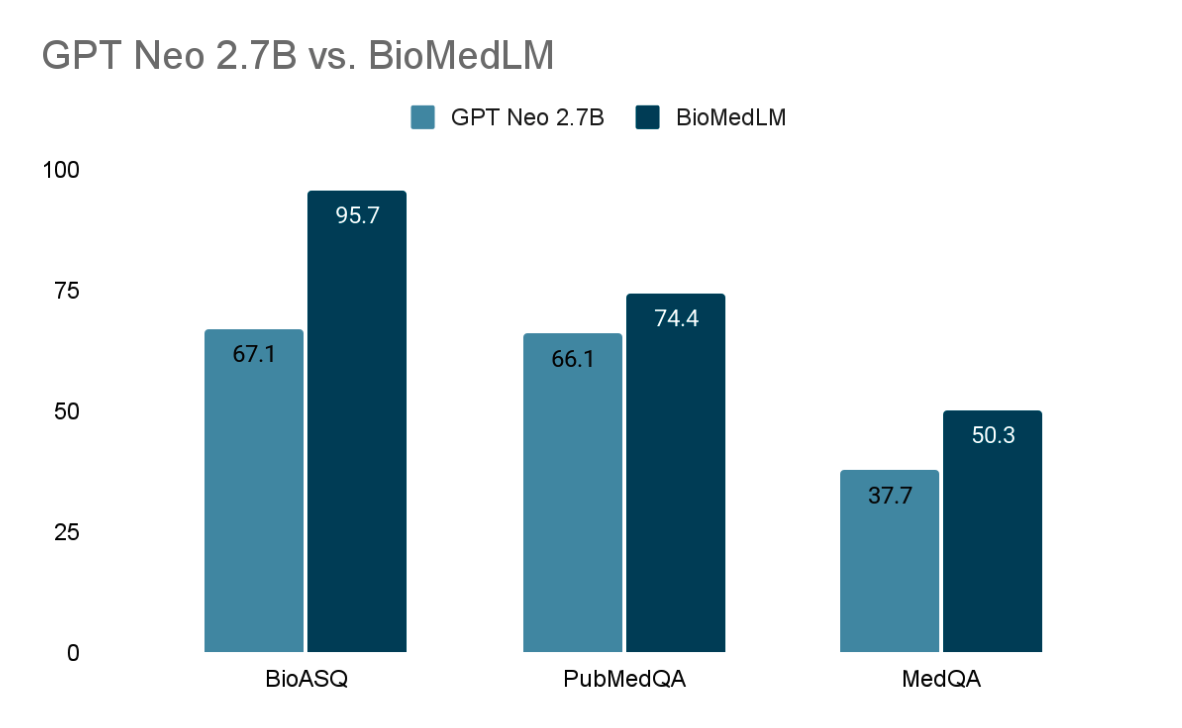}
\caption{\label{fig:biomedlm_vs_gpt_neo} Comparison of GPT-Neo 2.7B and BioMedLM on Select QA Tasks}
\end{figure}

Multiple works have demonstrated that domain-specific training can be beneficial for domain-specific downstream tasks \citep{legal}. To explore the extent to which pre-training on domain-specific tasks can help, we ran several experiments fine-tuning GPT-Neo 2.7B on biomedical QA tasks. GPT-Neo 2.7B is an architecturally similar model to BioMedLM with a nearly identical parameter count, but trained on general English. Figure~\ref{fig:biomedlm_vs_gpt_neo} shows the relative performance of the two systems on select tasks.

On each of the three select tasks, BioMedLM substantially outperforms GPT-Neo 2.7B, including a 27\% accuracy increase on BioASQ. Given that there can be substantial gains at the 2.7B scale by pre-training on PubMed for a long time, it suggests larger models could also benefit on these types of tasks if pre-trained for longer specifically on PubMed.

% \section{BioGPT Benchmarks}
% The BioGPT paper provides five benchmarks; three for relation extraction tasks, one for question-answering and a final one for document classification.

% \section{GNBR Dataset}
% The Global Network of Biomedical Relations (GNBR) is a network consisting of chemical (drug), disease and gene entities. We create a dataset that has a 150/150/150 train/validation/test set from the broader network. 

\section{Usage of BioMedLM since release}

Since its release in December 2022, BioMedLM has been evaluated on several biomedical benchmarks. It was evaluated on medical QA tasks (MedQA, PubMedQA) with accuracy 50.3\% and 74.4\%, respectively \citep{tian2023opportunities}. It has been used to improve the transferability of clinical note section classification models, using data such as discharge summaries, colorectal clinical notes and progress notes \citep{zhou2023transferability}. In \citet{yun2023appraising}, it was evaluated on text generation of medical systematic reviews after it was given the title of a review article. It has also been evaluated for general medical named entity recognition \citep{deusser2023informed} using bio-entity tasks JNLPBA and NCBI-Disease. BioMedLM performance was state of the art (with micro F1-score of 81.31\%), showing that pre-training on a specific domain helps with named entity decoding.

Other benchmarks involve relation extraction and classification, in various biomedical domains. For protein pathways and interactions \citep{park2023comparative}, tasks include recognizing protein-protein interactions (STRING Task 1 and 2), KEGG pathway recognition (KEGG Task 1 and 2) and evaluating gene regulatory relations (INDRA DB). Interestingly, in STRING Task 2 (classifying existence of any association or interaction between two proteins), BioMedLM demonstrated the most favorable performance (with micro F1-score of 0.643), while larger models, e.g., LLaMA, Alpaca, and BioGPTLarge, exhibited higher rates of false positives, possibly influencing a bias towards positive answers. 

In the microbiome field, BioMedLM was fine-tuned to extract microbiome-disease interactions \citep{karkera2023microbiome}. BioMedLM exhibited the best precision (0.822), outperforming GPT-3 (0.810). However, GPT-3 had the best accuracy of 0.814 vs.\ BioMedLM’s 0.806 and BioGPT's 0.732, possibly owing to its pre-training corpus of PubMed articles. In genetics, BioMedLM has also been evaluated on a benchmark, GeneTuring test, which focuses on nomenclature, genomic location, functional analysis and sequence alignment \citep{jin2023genegpt}. In the multi-modal space, there have also been efforts to use BioMedLM in vision-language pre-training models for computer-aided diagnosis \citep{chen2023medblip}; or prefix tuning for visual question answering tasks \citep{vansonsbeek2023openended}. 

In summary, given BioMedLM's medium size, as a public, low-cost model, it punches above its weight. There exists a tension between domain-specific knowledge either hindering or enhancing performance, where the former can be explained by over-optimization on medical text corpora, and the latter suggests the benefits of domain-specific information. BioMedLM's training on domain knowledge in PubMed lends it to better performance in biology-related tasks over clinical tasks. In relation extraction tasks, lack of extraneous out-of-domain data gives it a lower false positive rate and higher precision compared to other models.

\section{Conclusion}

We present BioMedLM, a 2.7 billion parameter GPT-style model trained on PubMed text. On biomedical question-answering tasks, BioMedLM can outperform bidirectional models with richer data sources and compete with the few-shot performance of models that have orders of magnitude more parameters. BioMedLM can also produce useful generations, delivering multi-sentence answers to medical knowledge questions. Our work demonstrates the potential of medium-sized models trained on domain-specific text.

\section{Acknowledgments}

We thank DataBricks (originally, MosaicML) for providing the compute for this project.

\section{Reproducibility}

The pre-trained model is available on the Hugging Face hub: 

\href{https://huggingface.co/stanford-crfm/BioMedLM}{https://huggingface.co/stanford-crfm/BioMedLM}

Code used for pre-training and fine-tuning is available on GitHub: 

\href{https://github.com/stanford-crfm/BioMedLM}{https://github.com/stanford-crfm/BioMedLM}

\newpage

\bibliographystyle{plainnat}
\bibliography{references}

\clearpage
\appendix
\counterwithin{figure}{section}
\counterwithin{table}{section}

\renewcommand{\thesubsection}{\arabic{subsection}}

\section{Examples Of Questions}
\setcounter{subsection}{0}

\hypertarget{medmcqa}{\subsection{MedMCQA Example Questions}}
\label{app:medmcqa}

\begin{table*}[h]
    \centering
    \begin{tabular}{p{7.5cm} p{7.2cm}}
     \toprule
     Question & BioMedLM Answer \\
     \midrule
     All of the following statements are true regarding hypertrophy, except: & {A) Occurs due to synthesis and assembly of additional intracellular components.
    
    B) There is an increase in the size of the cells.
    
    C) Cells capable of division respond to stress by hypertrophy and hyperplasia.
    
    \textbf{D) There is an increase in the number of cells.}} \\
    \midrule
    A 19 year old female presents with pain in the neck for 5 days. She is not able to wear tie for her job because of neck pain. H\/O fatigue and lethargy for 10 days. She had flu like symptoms 20 days ago which resolved spontaneously. BP 110\/80 mmHg, Pulse 48\/min. Extremities are cold and dry. Neck is very tender. ECG normal. TSH is elevated. ESR 30 mm\/hr. Next appropriate step & {A) Atropine injection
    
    B) Levothyroxine administration
    
    \textbf{C) Aspirin}
    
    D) Increase iodine intake in food} \\
    \bottomrule
    \end{tabular}
    \caption{Example Questions for MedMCQA}\label{tab:medmcqa-examples}
\end{table*}
\FloatBarrier

\hypertarget{medqa}{\subsection{MedQA Example Question}}
\label{app:medqa}

\begin{table*}[h]
    \centering
    \begin{tabular}{p{7.5cm} p{7.2cm}}
    \toprule
    Question & BioMedLM Answer \\
    \midrule
    A 45-year-old woman presents to the emergency department with acute onset of severe right upper quadrant abdominal pain that radiates to the infrascapular region. Her medical history is significant for obesity, hypertension, obstructive sleep apnea, and gastric bypass surgery 2 years ago after which she lost 79 kg (150 lb). The patient complains of nausea and vomiting that accompanies the pain. Her temperature is 38.9°C (101.2°F), blood pressure is 144/88 mm Hg, heart rate is 76/min, and respiratory rate is 14/min (fever). Abdominal examination is significant for right upper quadrant tenderness along with guarding and cessation of inspired breath on deep palpation of the right upper quadrant. Which test should be ordered first for this patient? & {\textbf{A) Abdominal ultrasound}
    
    B) CT scan of the abdomen
    
    C) MRI of the abdomen
    
    D) X-ray film of the abdomen} \\
    \bottomrule
    \end{tabular}
    \caption{Example Question for MedQA}\label{tab:medqa-examples}
\end{table*}
\FloatBarrier

\clearpage

\hypertarget{mmlu}{\subsection{MMLU Example Question}}
\label{app:mmlu}

\begin{table*}[h]
    \centering
    \begin{tabular}{p{7.5cm} p{7.2cm}}
     \toprule
     Question & BioMedLM Answer \\
     \midrule
     Zinc finger proteins and helix-turn-helix proteins are & {\textbf{A) types of DNA-binding proteins}
     
    B) involved in the control of translation
    
    C) components of ribosomes
    
    D) part of the hemoglobin in blood cells} \\
    \bottomrule
    \end{tabular}
    \caption{Example Question for MMLU}\label{tab:mmlu-examples}
\end{table*}
\FloatBarrier

\hypertarget{pubmedqa}{\subsection{PubMedQA Example Question}}
\label{app:pubmedqa}

\begin{table*}[h]
    \centering
    \begin{tabular}{p{15cm}}
    \toprule
    Context: \\
    To evaluate the degree to which histologic chorioamnionitis, a frequent finding in placentas submitted for histopathologic evaluation, correlates with clinical indicators of infection in the mother. A retrospective review was performed on 52 cases with a histologic diagnosis of acute chorioamnionitis from 2,051 deliveries at University Hospital, Newark, from January 2003 to July 2003. Third-trimester placentas without histologic chorioamnionitis (n = 52) served as controls. Cases and controls were selected sequentially. Maternal medical records were reviewed for indicators of maternal infection. Histologic chorioamnionitis was significantly associated with the usage of antibiotics (p = 0.0095) and a higher mean white blood cell count (p = 0.018). The presence of 1 or more clinical indicators was significantly associated with the presence of histologic chorioamnionitis (p = 0.019). \\
    \midrule
    Question: \\
    {Does histologic chorioamnionitis correspond to clinical chorioamnionitis? 
     
     \textbf{Yes}/No/Maybe}\\
    \bottomrule
    \end{tabular}
    \caption{Example Question for PubMedQA}\label{tab:pubmedqa-examples}
\end{table*}
\FloatBarrier

\clearpage

\hypertarget{bioasq}{\subsection{BioASQ Example Question}}
\label{app:bioasq}

\begin{table*}[h]
    \centering
    \begin{tabular}{p{15cm}}
     \toprule
     Context: \\
    NT-pro-BNP was significantly elevated postexercise in both adults and adolescents and remained above baseline at 24 h in both groups. NT-pro-BNP concentrations increased significantly (28 +/- 17.1 vs 795 +/- 823 ng x L, P < 0.05), whereas postrace cTnT were elevated in just five athletes (20\%). [NT-pro-BNP] was observed immediately after the marathon (median [NT-pro-BNP] before: 39.6 pg ml(-1), after: 138.6 pg ml(-1), p=0.003) with a further increase on day one. [BNP] did not increase immediately after the marathon but increased on day one (median [BNP] before: 15 pg ml(-1), day one: 27.35 pg ml(-1), p=0.006). Pro-BNP was significantly increased immediately post-race (27+/-21 vs 7+/-2 pmol/L pre-race, P < or = 0.007), which 12-24 h later, decreased to 19+/-14 pmol/L (P = 0.07 vs pre-race). The relatively high NT-proBNP levels after active recovery when psychophysical stress is higher, because of cycling and cold water immersion, suggest that not only endurance exercise, but also strenuous, stressful short exercise can induce an increase in NT-proBNP concentrations. Running a marathon significantly increases NT-pro-BNP levels in healthy adults. This increase could be partially attributed to cardiac stress. Increases in NT-proBNP can be found in a major part of obviously healthy athletes after prolonged strenuous exercise. The release of BNP during and after exercise may not result from myocardial damage but may have cytoprotective and growth-regulating effects. The different nature of exercise-induced increases in BNP and cardiac troponins has to be elucidated in the future. In healthy cyclists, transient increases in NT-pro-BNP and cTnT are more likely to reflect cardiac fatigue than injury. The rise in BNP in older athletes may reflect a reversible, mainly diastolic left ventricular dysfunction.  Plasma BNP concentrations were higher in both the judo and marathon groups than in controls, and positively correlated with LV mass as well as with deceleration time. Such exercise significantly increased ANP and BNP levels in healthy men, and the increases could be partially attributed to myocardial damage during the race. \\
    \midrule
    Question: \\
    {Does BNP increase after intensive exercise in athletes?

    \textbf{Yes}/No} \\
    \bottomrule
    \end{tabular}
    \caption{Example Question for BioASQ}\label{tab:bioasq-examples}
\end{table*}
\FloatBarrier

\end{document}